\documentclass[12pt]{article}
\usepackage{titlesec}
\usepackage{graphicx} 
\usepackage[margin=1in]{geometry} 
\usepackage{hyperref}

\title{Pre-Trained CNN Architecture for Transformer-Based Image Caption Generation Model}
\author{\footnotesize Dufera Amanuel Tafese \\ \footnotesize Xi'an Jiaotong University}

\date{} 

\begin{document}

\maketitle

\section*{Abstract}
Automatic image captioning, a multifaceted task bridging computer vision and natural language processing, aims to generate descriptive textual content from visual input. While Convolutional Neural Networks (CNNs) and Long Short-Term Memory (LSTM) networks have achieved significant advancements, they present limitations. The inherent sequential nature of RNNs leads to sluggish training and inference times. LSTMs further struggle with retaining information from earlier sequence elements when dealing with very long sequences. This project presents a comprehensive guide to constructing and comprehending transformer models for image captioning. Transformers employ self-attention mechanisms, capturing both short- and long-range dependencies within the data. This facilitates efficient parallelization during both training and inference phases. We leverage the well-established Transformer architecture, recognized for its effectiveness in managing sequential data, and present a meticulous methodology. Utilizing the Flickr30k dataset, we conduct data pre-processing, construct a model architecture that integrates an EfficientNetB0 CNN for feature extraction, and train the model with attention mechanisms incorporated. Our approach exemplifies the utilization of parallelization for efficient training and inference.\\
You can find the project on \href{https://github.com/amani-td/Caption-Generation}{GitHub}.

\section{Introduction}
Many of the captioning algorithms now in use to capture the essence of an image in words are based on the encoder-decoder architecture, where a decoder infrastructure can use an attention method to predict words by employing a function obtained from an encoder network. The translation strategy using a visual encoder and a language decoder has been the major focus of studies on picture subtitling \cite{Subash_2019}. 

Image captioning serves many purposes, ranging from enabling autonomous driving and enhancing face recognition systems to assisting individuals with visual impairments and improving photo search quality. This challenging task involves extracting image features through computer vision (CV) techniques and generating natural language descriptions using natural language processing (NLP) technology. The impact of photo caption generation extends across diverse domains, including image search, software development for accessibility, video surveillance, security, and human-computer interaction \cite{10.1145/3295748}.

The challenge of generating photo captions, a prominent task in sequence modeling, has seen the adoption of various methodologies. One popular approach combines Convolutional Neural Networks (CNNs) with Recurrent Neural Networks (RNNs) in a CNN-RNN framework. However, training RNN-based models, especially those like Long Short-Term Memory (LSTM) networks, can be challenging due to their sequential nature, leading to suboptimal performance \cite{9182105}

To address this, recent advancements have introduced the Transformer architecture, overcoming the limitations of sequential processing through its parallelizable design and context-aware attention mechanism. Unlike traditional RNN-based models, Transformers can operate efficiently during both training and inference without relying on specific sequences, thus achieving state-of-the-art results in photo captioning tasks \cite{Vaswani2017AttentionIA} \cite{10.1145/3115432}

The focus of caption generation studies has primarily been on English-language datasets, leading to the widespread adoption of attention-based mechanisms. While many utilize the VGG-16 architecture as the encoder in captioning models, alternative approaches incorporate pre-trained networks like AlexNet or Residual Networks (ResNet) to extract visual features. Additionally, some researchers leverage Bidirectional Long Short-Term Memory (BiLSTM) networks for language modeling in caption generation tasks\cite{Aneja_2018_CVPR} \cite{10.1145/3115432}

In addition to English, various datasets have been curated for caption generation in other languages such as Chinese, Japanese (e.g., Yoshikawa dataset), Arabic, and Bahasa Indonesia. Custom datasets have also been developed, combining sources like MS COCO and Flickr30k for languages like Indonesian. Moreover, specific language versions of popular datasets like Flickr30k, such as FEEH-ID Flickr8k, have been created to cater to diverse linguistic needs \cite{Almuzaini2018AutomaticAI} \cite{yoshikawa2017stair}\cite{10.1145/3123266.3123366}

\cite{Zhu2018CaptioningTW} introduced a novel approach by leveraging Transformers in two distinct streams within their architecture: one for the visual component and another for the linguistic aspect. They combined a CNN model for encoding image features with a Transformer model for decoding captions. This architecture involved separate Transformer models for both the encoder and decoder, incorporating stacked self-attention mechanisms. \cite{8787715} also demonstrated the effectiveness of using CNN as an encoder to extract image features, generating a context vector that encapsulates essential visual information. This vector is then input into the Transformer for caption generation based on the extracted features.

The structure of this paper is outlined as follows: In Section 3, reviews the related work in the field. Section 4 provides an overview of the methodology of this study and section 5 details of our model design and implementation, while Section 6 offers a detailed performance analysis, highlighting the impact of different configurations. Finally, in Section 7, we draw conclusions.

\section{Related works}
Image description is an important problem in the field of Artificial Intelligence, in simple terms, the input is an image to the model, and the output of the model is a text sentence that can describe the image scene. There are many methods related to image description, the more famous ones are:\\
	Traditional methods: early approaches to image caption generation typically used a combination of Convolutional Neural Networks (CNNs) and Long Short-Term Memory Networks (LSTMs) or Recurrent Neural Networks (RNNs). These methods encode images into feature vectors and then use RNN or LSTM to generate subtitles\cite{10.1145/3617592}.\\
	Attention Mechanism Based Methods: later methods introduced an attention mechanism that allows the model to focus on different regions in the image when generating subtitles, thus improving performance and generation quality \cite{LIU2020102178}.\\
	Transformer-based approaches: Recently, the Transformer architecture has achieved great success in natural language processing tasks, and some research has begun to apply it to image caption generation tasks. These approaches use Transformer to capture complex relationships between images and text to generate more accurate and coherent captions\cite{app131911103}.
When talking about work related to image caption generation, some classic papers and approaches are a must see, they give inspiring thoughts and give us a deeper understanding of the field, here are some of the classic papers as well as some of the latest technological advances.\\
	Show and Tell: A Neural Image Caption Generator \cite{vinyals2015tell}:
This is one of the classic work in the field of image caption generation. It introduces an end-to-end deep learning model that combines CNN and LSTM for encoding images into feature vectors and generating corresponding captions. The method was one of the pioneers in the task of image caption generation and laid the foundation for subsequent research.\\
	Show, Attend and Tell: Neural Image Caption Generation with Visual Attention \cite{xu2016show}:
This work introduces an attention mechanism based on the classical Show and Tell model, allowing the model to dynamically focus on different regions in the image when generating subtitles. In this way, the model can more accurately describe the image content and improve the quality of the generated subtitles.\\
	Bottom-Up and Top-Down Attention for Image Captioning and Visual Question Answering \cite{Anderson2017BottomUpAT}
This work proposes a combined approach of bottom and top-level attention mechanisms for image caption generation and visual quizzing tasks. The bottom-level attention mechanism extracts local information from the image, while the top-level attention mechanism integrates this information in a global context. This combination allows the model to better understand images and generate more coherent captions.\\
	BERT (Bidirectional Encoder Representations from Transformers) \cite{devlin2019bert}
BERT is a pre-trained Transformer-based language model that has achieved great success in the field of natural language processing. While originally designed to process textual data, some research has also begun to explore the application of BERT or similar Transformer models to the task of image caption generation to capture the complex relationships between images and text.\\
	Image Transformer \cite{parmar2018image}
This work proposes a pure Transformer architecture for processing image data. It partitions an image into a series of blocks and processes these blocks through the Transformer model, which enables the learning of a representation of the image. Although this approach is initially used for image classification and segmentation tasks, it also provides a new idea for applying Transformer to image related tasks.
These papers represent some of the classic and most recent work in the field of image caption generation, demonstrating advances in deep learning and natural language processing in the field. The language in which the captions are ultimately generated depends on the configured dataset, e.g., the processing of a language involves tasks such as word segmentation, speech recognition, machine translation, etc. Previous work may include building databases for that language, developing NLP tools and models based on that language, etc.

\section{Methodology}
The methodology adopted for this project incorporates experimenting of data on the deep learning model architecture. The Flicker30k datasets for image captions are available publicly, hence two procedures are followed for developing the image captioning system. The preparation of the dataset comes first, then comes the design and implementation of the model architecture.
Dataset Preparation:
 Two tasks are involved in the production of the dataset: preprocessing and dataset collecting. The resulting dataset is split into three sections: one for testing, one for validation, and one for training.
Dataset Collection:
The Flickr30k public dataset provided the datasets gathered for this study. It comprises of more than 300,00 photos with 5 captions each. The public datasets are available for download.

\begin{figure}
  \centering
  \includegraphics[width=\textwidth]{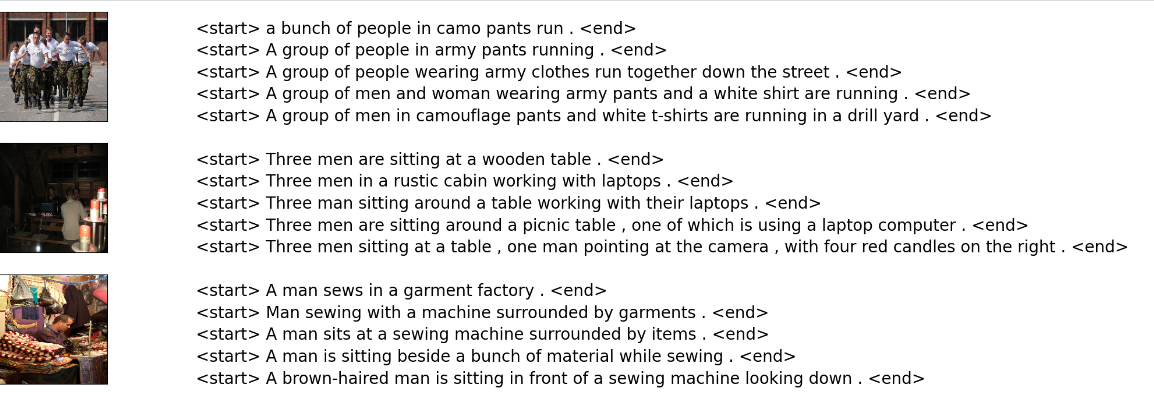} 
  \caption{An examples from the dataset: Images and the five captions of each image}
  \label{fig:enter-label}
\end{figure}

 Dataset Preprocessing:
Pre-processing involves caption normalization, length filtering, and dataset partitioning. Image processing operations, such as decoding, and resizing, are implemented. The dataset is prepared using TensorFlow functions, including shuffling, mapping, batching, and prefetching for optimal parallelization.
Dataset Creation: 
Following cleaning, the caption data are divided into three sets: a training set, a validation set, and a testing set. These sets contain, respectively, 20915, 5124, and 105 images. Using the TensorFlow "Dataset" library, the caption data are then mapped to the corresponding images and compressed to produce datasets for training and validation.
\section{Model Design and Implementation}
The suggested method for image captioning incorporates a CNN model for image feature extraction and a Transformer network for language modelling. Figure 2 displays the system architecture for this work. The CNN model can be developed from scratch but due to its advancement in recent years more accurate and efficient pre-trained models are available for use. Hence, ‘EfficientNetB0’ used to extract picture characteristics. The input image is encoded by CNN into a vector representation, which the decoder uses to produce captions. Since this job does not include categorization, the CNN's final softmax layer is eliminated
\begin{figure}
  \centering
  \includegraphics[width=0.9\textwidth]{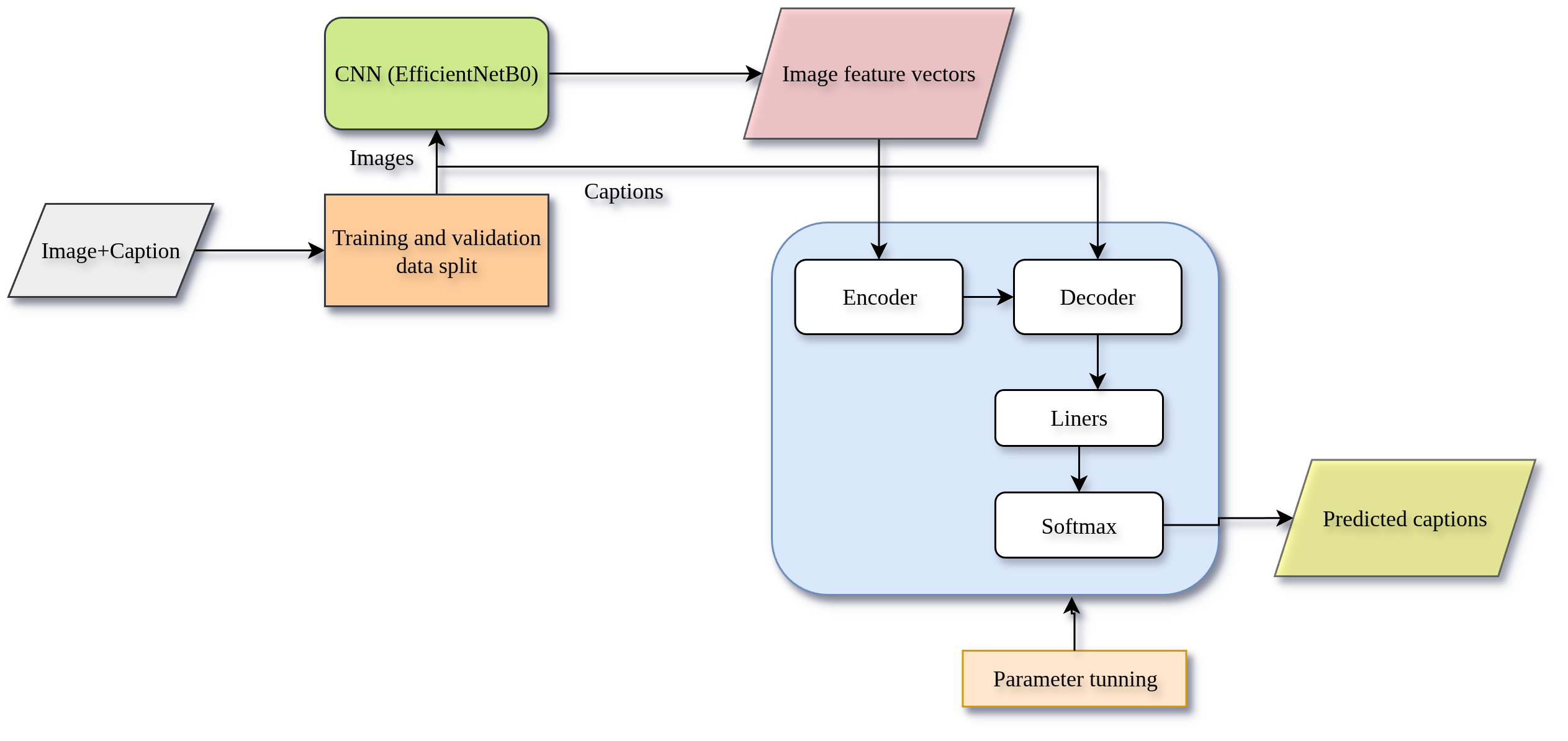}  
  \caption{The model architecture}
  \label{fig:enter-label}
\end{figure}

EfficientNet is a CNN model as well as a scaling technique that uses a set of preset scaling coefficients to uniformly scale depth, width, and resolution dimensions, in contrast to standard practice, which scales these variables arbitrarily \cite{tan2020efficientnet}
There are 8 variants of EfficientNet (B0 - B7). EfficeintNetB0 is used for this project because of its simplicity and relatively good performance. The transformer model is used to generate captions instead of the conventional Recurrent Neural Network-based architecture because RNN doesn’t support parallelization and transformer networks have outperformed RNN in NLP tasks in recent years.Our proposed system follows the transformer architecture proposed by \cite{Vaswani2017AttentionIA}, The main components of the transformer networks consist of the encoder, decoder, positional encoding, embeddings, softmax, and multiheaded attention. Attention is utilized in the transformer model to find the relevant collection of values based on a few keys and queries. Attention weights, which are derived using the encoder hidden state (Key) and decoder hidden state (Value), have recently been used to give priority to distinct encoder hidden states (values) in processing the decoder states (query) \cite{biology10050371}.

\begin{table}[]
    \centering
    \begin{tabular}{|c|c|}
    \hline
        Parameters &  Value \\
        \hline
       Image Size  & (299,299)\\
       Max. Vocab Size & 13000\\
       Sequence Length & 24\\
       Embedding Size & 512\\
      Batch Size & 128\\
      Optimizer& Adam\\
     Loss Function & Categorical-crossentropy\\
     \hline
    \end{tabular}
    \caption{Model Parameters}
    \label{tab:my_label}
\end{table}

Single attention-weighted values have been found to be insufficient to capture the many features of the input. The transformer model thus employs multiheaded attention for tackling this challenge. Similarly, positional encoding is used by the transformer networks to keep information about the order of sequence by adding the relative or absolute position of the tokens in sequence \cite{Vaswani2017AttentionIA}. The positional encodings are added to the bottom of the encoder and decoder stacks. In order to implement our model, the datasets generated are passed to both of the CNNTransformer architecture. The input images of size (299x299) are passed to the CNN encoder to generate image vectors. The image vectors are then passed to the transformer encoder. The transformer decoder part is fed with the respective captions to train the model. The encoder part comprises a single multi-headed attention head and a normalization layer whereas 2 multi-headed attention heads and 3 normalization layers are used in the decoder. These models are implemented using the TensorFlow Keras library. Table 1 shows the model parameters used in this work. The model parameters are chosen based on explicit experimentation on Flicker30K dataset. These are the optimal parameter values as per our research which can be further improved with the introduction of larger datasets and via parameter tuning.

\section{Results and Discussions}
In order to create captions, we employed Adam's optimization algorithm to train a Transformer-based model. This model was trained with a 30-epoch cross-entropy loss on our Flickr30k dataset. Prior to this training procedure, we utilized transfer learning by including components of a pre-trained model into our new model. Using a pre-trained encoder for photo captioning is usually better than starting from scratch with a fresh model and no prior knowledge. It is done within fifty epochs and halted if there has been no progress in BLEU-4 over the most recent 10 epochs (the stopping training criterion), while the total training procedure is done in 5-12 hours on each pre-trained CNN architecture. Python with Tensorflow's library is the performance analysis environment for each CNN model that contains three phases: (1) Training period. (2) Validation. (3) Testing. In other words, we apply the parallelization to it, as the Transformer’s design permits the simultaneous process.

A dataset of over 30,000 image-caption pairings is developed which are separated into training, validation and testing sets. The "test" results were assessed using our evaluation approach, where each sentence created by the model in the provided picture was compared to the target sentences used as sentence references to calculate the BLEU score. These scores measure only matching grams of a given order, such as single words (1-gram) or word pairs (2-gram or bigram), and so forth. BLEU score goes from 0 to 1 where a score between 0.6 to 0.7 is believed to be the greatest attainable outcome while at the same time, a score between 0.3 to 0.4 is considered an obviously decent translation and a score more than 0.4 is called high-quality translation(Google, 2022). In this work, BLEU score is generated on the complete test data at once using the NLTK bleu library. 
\begin{figure}
  \centering
  \includegraphics[width=0.85\textwidth]{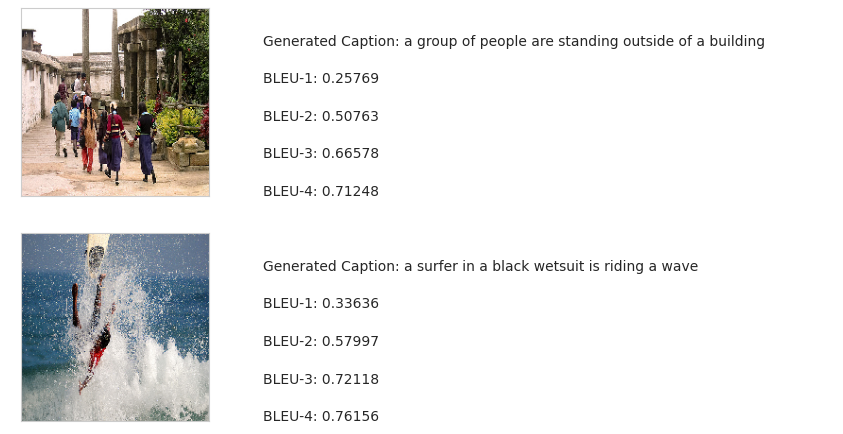}  
  \caption{Model output that shows the prediction caption with score of BLEU.}
  \label{fig:enter-label}
\end{figure}

Figure 3 shows the qualitative outcomes of the Transformer-based model's inference, as predicted directly. The picture does not exist in the prior training dataset, where specific images and words from our Transformer-based model create grammatically valid descriptions. On the other hand, the first two BLEU ratings are not as very good, but may nonetheless serve as a reference for picture captioning. It is also observed that the final two BLEU scores are greater than the first two values.
\begin{figure}
    \centering
    \includegraphics{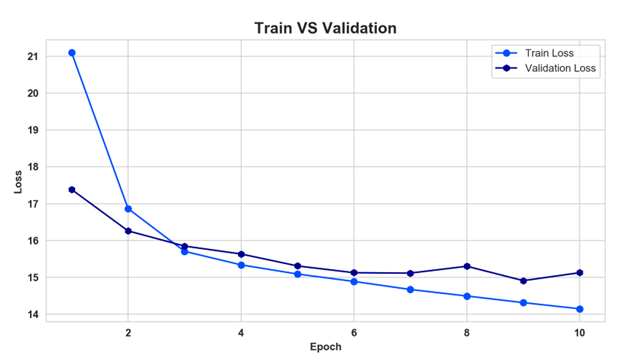}
    \caption{ The difference between Training loss and validation loss}
    \label{fig:enter-label}
\end{figure}
As show figure 4, validation data is used during model development to guide training and make decisions about the model. On the other hand, the accuracy of the model increasing with each epoich training.
\section{Conclusion}
This study focuses on the utilization of a Transformer model for image captioning tasks on a small dataset, offering a comprehensive resource for understanding and building Transformer models. The Transformer architecture, known for its efficiency in handling sequential data, particularly in NLP tasks, is explored. Key components such as input embedding, positional encoding, multi-head attention, and masking are explained in detail.

The methodology involves data preprocessing using the Flickr30k dataset, including caption normalization and image processing functions. The model architecture consists of a pre-trained EfficientNetB0 CNN for image feature extraction, followed by an encoder-decoder setup incorporating multi-head attention and layer normalization. Training involves updating weights to minimize loss, with the model evaluated using sparse categorical cross-entropy loss during validation and the Greedy algorithm combined with BLEU score for caption generation during inference.

Overall, this research contributes to the understanding and implementation of Transformer-based models for image captioning, offering insights into the architecture, training process, and evaluation techniques. The approach presented in this paper lays the foundation for further advancements in image captioning tasks using Transformer models on small datasets.

\bibliographystyle{plain}
\bibliography{11.bib,12.bib,13.bib,14.bib,15.bib,16.bib,17.bib,1.bib,2.bib,3.bib,4.bib,5.bib,6.bib,18.bib,19.bib,20.bib,21.bib,22.bib,23.bib,24.bib}

\end{document}